# An Uncertainty Management Calculus for Ordering Searches in Distributed Dynamic Databases*

Uttam Mukhopadhyay[†]
Center for Machine Intelligence, University of South Carolina,
Columbia, SC 29208

## Abstract

*MINDS is a distributed system of cooperating query engines that customizes document retrieval for each user in a dynamic environment. It improves its performance and adapts to changing patterns of document distribution by observing system-user interactions and modifying the appropriate certainty factors, which act as search control parameters. It is argued here that the uncertainty management calculus must account for temporal precedence, reliability of evidence, degree of support for a proposition, and saturation effects. The calculus presented here possesses these features. Some results obtained with this scheme are discussed.*

## 1 Introduction

Documents are used in computerized office environments to store a variety of information. This information is often difficult to utilize because:

1. Users do not have perfect knowledge of the documents in the system or the organization for their storage.

2. Document names and keywords are weak descriptors of document content: traditional search techniques based on keywords lack precision (defined as the percentage of retrieved documents that are relevant).

3. Document distribution patterns are dynamic in nature:

   - outdated documents are deleted
   - new documents are added
   - copies of existing documents are made (to be stored elsewhere)
   - documents are relocated, either *directly* or *indirectly* (by copying to another location and later deleting the original).

4. The interests, preferences and responsibilities of each user changes over time. One effect is that the relevance of a given document to a user may change over time. The composition of his locally stored documents may also change to mirror his changing interests.

MINDS (Multiple Intelligent Node Document Servers) is a distributed system of intelligent query engines that dynamically learn document storage patterns from the perspective of individual users so that searches are efficient and produce relevant documents [1],[2].

## 2 Query Environment

Queries for documents may be predicated on a number of attributes such as document name, author, creation date, etc. Queries based on document content present some special research issues, because simple descriptors, like keywords, do not adequately describe document content. More complex descriptors might alleviate the problem but they are impractical in most office environments because a lot of effort is required to describe documents and issue queries, which would necessarily be more complex in order to exploit the richer descriptions of the documents. A detailed description of the dynamic profiles of the users would also be required by such a system.

The objective here is to retain the simplicity of a keyword-based system and yet achieve the performance levels of a more sophisticated system. This is enabled by the heuristics that effectively manage

*This research was supported in part by NCR Corporation

[†]Present address: Computer Science Department, General Motors Research Laboratories, Warren, MI 48090

the uncertainties associated with plausible search sequences.

In the environment considered here, each user owns a set of documents; each document is described by a few keywords. When a user issues a request for documents with a specific property (set of keywords), the system conducts a search of each set of documents and arranges the returned subsets from the most relevant to the least relevant.

With $n$ users systemwide, there will be $n$ returned subsets of documents and $n$-factorial possible orderings of the document subsets for each query. The order chosen by the system is based on the certainty factors that are stored and continuously updated by the system while executing the queries. The certainty factor relating two users and a keyword is a measure of the belief that the first user has of the ability of the set of documents owned by the second user to provide relevant knowledge in the area described by the keyword. An example of a heuristic for modifying certainty factors is:

> If Smith issues a command for reading documents with the keyword 'compilers' and at least one document is returned from Brown's set of documents, then
>
> 1. update the certainty factor that Smith associates with Brown's documents on 'compilers'
>    - the degree of support is equal to the maximum of all the relevance values accorded by Smith to Brown's documents on 'compilers'
>    - the reliability of this evidence is 1.0 (the maximum)
> 2. update the certainty factor that Brown associates with Smith's documents on 'compilers'
>    - the degree of support is 1.0
>    - the reliability of this evidence is 0.1 (Brown assumes that Smith may acquire new documents on 'compilers' that may be relevant to him)

## 3 Managing Uncertainty

### 3.1 Constraints

It is argued that the uncertainty management calculus should be able to take into account the following:

1. **Temporal Precedence**—the system is dynamic and therefore recently acquired evidence is more indicative of the current state of the system than evidence gathered earlier. If $f1$ is the mapping function for a downward revision of the certainty factor (contradiction) and $f2$ is the mapping function for an upward revision (confirmation), then:

   $$f2(f1(x)) \geq f1(f2(x))$$

   for all $0 \leq x \leq 1$ (Figure 1)

2. **Reliability of Evidence**—some types of evidence are more reliable than others. If a document with a desired keyword is successfully retrieved by Jones from Smith, this action by itself does not completely support the proposition that Smith's documents on compilers will be relevant to Jones in the future, since the relevance of this document to Jones is not known. However, if this document is read by Jones, then the relevance value assigned by him constitutes reliable evidence.

3. **Degree of Support**—the degree of support (membership) for a proposition may vary. When a user is asked by the system if the document he has just read is relevant to him, his answer does not have to be limited to 'yes' or 'no' but may be something in between, indicated on a range of numbers. (Incidentally, this type of evidence has very high reliability).

4. **Saturation Characteristics**—when the initial certainty factor for a metaknowledge element is high, additional confirmatory evidence will not change (increase) it substantially. However, if the evidence were to be contradictory, the change (decrease) in confidence factor would be high under the same initial condition. The situation is exactly reversed when the initial certainty factor is low.

### 3.2 Proposed Calculus

The uncertainty management calculus presented here has all these features and is based on two functions that map the current certainty factor to a new one (Figure 1). The function $f2$ deals with confirmatory evidence that causes upward revision while $f1$ deals with contradictory evidence that causes downward revision. A large set of function pairs would satisfy the requirements outlined above. Additional

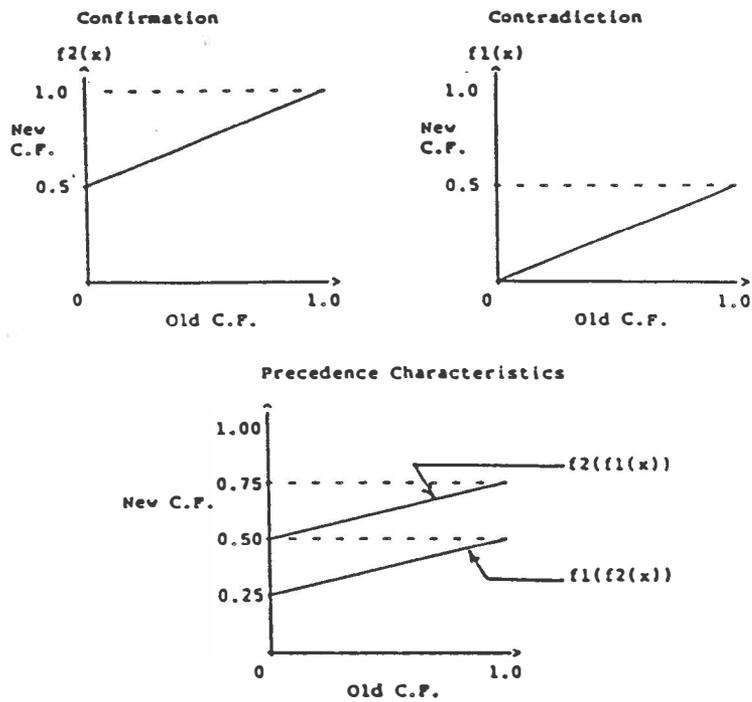

FIG. 1 Update functions for metaknowledge certainty factor and temporal precedence characteristics.

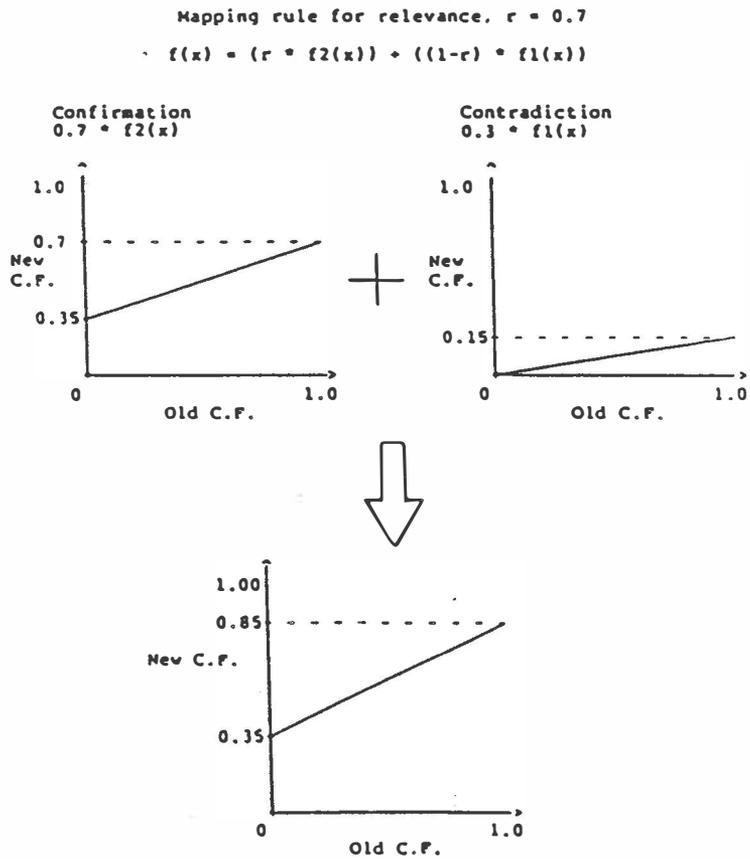

FIG 2 Updating scheme for metaknowledge certainty factors.

intuitively appealing constraints may be used to prune the space of candidate functions:

- $f1$ must decrease monotonically
- $f2$ must increase monotonically
- $f1(x) \leq x$ for all $0 \leq x \leq 1$
- $f2(x) \geq x$ for all $0 \leq x \leq 1$
- there is an upper bound on the new certainty factor when the most recent evidence contradicts the implied proposition
- there is a lower bound on the new certainty factor when the most recent evidence confirms the implied proposition.

The set of functions $f1$ and $f2$ shown in Figure 1 satisfies all these constraints.

When a keyword-based retrieval is executed, the revised certainty factor is given by:

$$(1-q) * x + q * (f(x))$$

where $x$ is the original certainty factor, $q$ is the reliability of this type of evidence (in the range $[0,1]$) and $f$ is the updating function: if the evidence supports the proposition (implied in the metaknowledge tuple) to a degree of 0.7, then $f$ is the weighted average of $f2$ and $f1$ in the ratio 0.7 : 0.3 (Figure 2).

It is important to note that this framework allows some leeway in tuning the calculus for trading off noise immunity against fast response times. Also, for a given calculus, two degrees of freedom (reliability of evidence and degree of support) are available for tuning each heuristic.

## 4  Results and Future Research

Some results of a system simulation are shown in Figure 3. The system improves its performance over time. The complete set of certainty factors represents a form of metaknowledge: the knowledge about the distribution pattern of object-level knowledge (contained in documents). The dissimilarity between the actual metaknowledge and the ideal metaknowledge provides a measure for system performance. A heuristic distance measure is used for this purpose—smaller distances imply improved performance.

The heuristics and the calculus are closely intertwined so that an independent analysis of either one requires extensive experimentation. Other competing uncertainty management calculi for dynamic environments need to be investigated.

## 5  Acknowledgement

I would like to thank Mike Huhns of MCC, Austin, TX and Ron Bonnell and Larry Stephens of the Center for Machine Intelligence, University of South Carolina, for lending me their *minds*.

## References

[1] U. Mukhopadhyay, L. M. Stephens, M. N. Huhns, and R. D. Bonnell, 'An Intelligent System for Document Retrieval in Distributed Office Environments' *Journal of the American Society for Information Science*, May 1986, Volume 37, Number 3, 123–135.

[2] R. D. Bonnell, M. N. Huhns, L.M. Stephens, and U. Mukhopadhyay, 'MINDS: Multiple Intelligent Node Document Servers' *Proceedings IEEE First International Conference on Office Automation*, December 1984, 125–136.

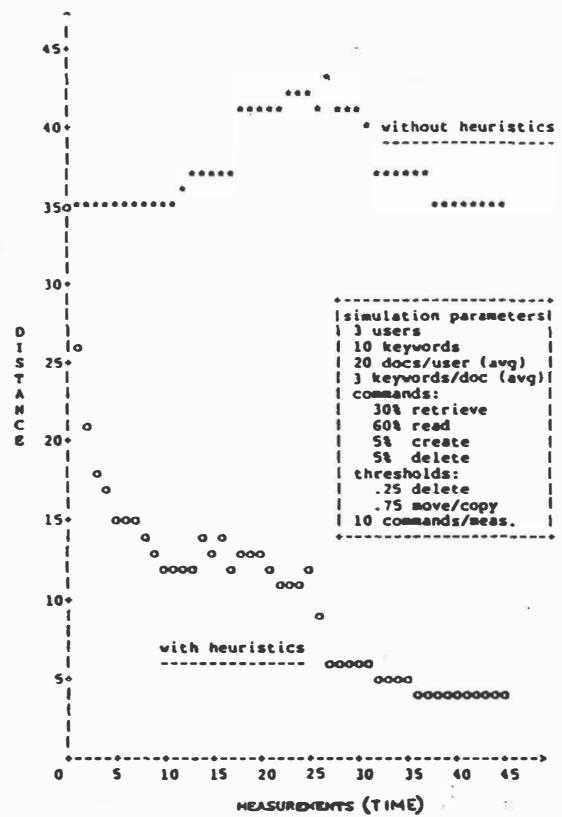

FIGURE 3  Learning Curve I (Distance Between Actual and Ideal Metaknowledge Decreases With System Usage)